%% file: main.tex
\documentclass[acmsmall,nonacm]{acmart}

\setcopyright{none}
\usepackage{amsmath}
\usepackage{algorithm}
\usepackage{algpseudocode}
\usepackage{graphicx}
\usepackage{textcomp}
\usepackage{xcolor}
\usepackage{listings}
\usepackage{enumitem}

\begin{document}

\title{CiteRadar: A Citation Intelligence Platform for Researcher Profiling and Geographic Visualization}

\input{sigconf_dochead}
\include{sigconf_author}

\maketitle

\fancyhf{}
\fancyfoot[C]{\thepage}
\renewcommand{\headrulewidth}{0pt}
\renewcommand{\footrulewidth}{0pt}

\input{1_introduction}
\input{2_background}
\input{3_newmethod}
\input{4_evaluation}
\input{5_casestudy}
\input{6_relatedwork}
\input{7_conclusion}

\input{sigconf_ack}

\input{sigconf_bib}

\end{document}

%% file: sigconf_dochead.tex
\input{0_abstract}


%% file: 0_abstract.tex
\begin{abstract}

Understanding the geographic reach and community structure of one's scholarly
citations is increasingly valuable for career development, grant applications, and collaboration discovery—yet accessible tools for answering these questions remain scarce. Existing bibliometric platforms either require costly institutional subscriptions or expose only aggregate citation counts without granular per-author metadata.

We present \textbf{CiteRadar}, an open-source system that accepts a
single Google Scholar user identifier and automatically produces a structured
output folder containing: the author's complete publication list, all
retrieved citing papers with enriched author metadata, two ranked author
tables (by citation frequency and by h-index), a plain-text statistical
summary, and a self-contained interactive HTML world map—all from a single
command-line invocation. CiteRadar integrates five heterogeneous data sources—Google Scholar, OpenAlex, CrossRef, Semantic Scholar, and OpenStreetMap Nominatim—through
a carefully engineered five-stage pipeline.
Key technical contributions include:
(1) a Scholar meta-string parser resilient to Unicode non-breaking-space
    separators, a pervasive but undocumented quirk in Scholar's HTML that
    silently corrupts venue and year fields when unhandled;
(2) a two-stage author disambiguation system using stop-word-filtered
    institution name similarity to guard against the well-known same-name
    entity-merging failure mode in bibliometric databases, demonstrated to
    eliminate h-index attribution errors of up to $9\times$ the correct value;
(3) an OpenAlex web-URL to API-URL conversion fix that raises the fraction
    of author records with city-level location data from 0\% to $\approx$60\%;
and (4) a logarithmically-scaled interactive Folium world map with per-city
    researcher popups, rendered as a fully self-contained HTML file. CiteRadar is available at \url{https://github.com/chenxuniu/citeradar}
and installable via \texttt{pip install citeradar}.

\end{abstract}

%% file: sigconf_author.tex
\author{Chenxu Niu}
\affiliation{%
  \institution{NVIDIA Corporation}
  \city{Santa Clara}
  \state{California}
  \postcode{95051}
}
\email{chenxun@nvidia.com}

\author{Yiming Sun }
\affiliation{%
  \institution{Texas Tech University}
  \city{Lubbock}
  \state{Texas}
  \postcode{79409}
  }
\email{yiming.sun@ttu.edu}


%% file: 1_introduction.tex
\section{Introduction}
\label{sec:intro}

The impact of a research publication is traditionally measured through aggregate bibliometric indicators: total citation count,
h-index~\cite{hirsch2005index}, journal impact factor, and field-normalized metrics. These numbers answer \textit{how much} a body of work has been cited, but they systematically obscure a richer and more actionable set of questions
that individual researchers increasingly care about:

\begin{itemize}[leftmargin=*, itemsep=2pt]
  \item \textit{Who specifically} is building on this work, what are their full names and institutional affiliations?
  \item \textit{Where} in the world are these researchers located, and does the geographic distribution reveal clusters of concentrated interest?
  \item \textit{How influential} are the citing researchers themselves? Do citations come from established leaders or primarily from early-career authors?
  \item Are there natural collaboration candidates already demonstrably familiar with this research direction?
\end{itemize}

Answers to these questions carry concrete practical value. A researcher preparing a grant proposal can cite specific institutions and countries where independent groups have adopted their methodology, providing evidence of genuine community uptake. A junior faculty member building a promotion case can identify established senior scholars (verifiable by name, institution, and h-index) whose citations validate the significance of the work. A scientist seeking new collaborators can prioritize outreach to groups already familiar with the foundational methodology.

Despite this demand, accessible tools for answering these questions are scarce. Commercial platforms such as Web of Science and Scopus offer citing-author
metadata but require institutional subscriptions costing tens of thousands of dollars annually. Google Scholar~\cite{vine2006google} is the most widely used free academic search engine, indexing over 350 million documents including preprints and grey literature, but its public interface exposes only aggregate citation
counts per paper without structured per-author affiliation metadata. Open bibliometric APIs such as OpenAlex~\cite{priem2022openalex} and CrossRef~\cite{hendricks2020crossref} provide powerful programmatic access but require significant bespoke engineering to assemble into an end-to-end tool.

We present \textbf{CiteRadar}, a Python command-line tool that bridges this gap. Given a Scholar user ID (the 12-character code embedded in any Scholar profile URL), CiteRadar executes a fully automated five-stage pipeline: (1)~scraping the complete publication list; (2)~following every ``Cited by'' link to collect all citing papers with author-list enrichment via CrossRef; (3)~resolving full per-author metadata through a priority cascade of OpenAlex, Semantic Scholar, and CrossRef; (4)~building two ranked author tables using a novel two-stage disambiguation algorithm; and (5)~generating a statistical summary and an interactive world map.

The entire analysis requires a single command:
\begin{lstlisting}[language=bash, numbers=none]
pip install citeradar
citeradar YOUR_SCHOLAR_ID --outdir your_folder
\end{lstlisting}

All outputs are deposited in a folder named after the researcher name, making each analysis self-contained and shareable.

%% file: 2_background.tex
\section{Related Work}
\label{sec:related}

\subsection{Commercial Bibliometric Platforms}

Web of Science (Clarivate)~\cite{webofscience2026} and Scopus (Elsevier)~\cite{scopus2026} both provide citing-author affiliation data through analytical interfaces and APIs. Both require institutional subscriptions, effectively limiting access to researchers at well-resourced universities. Neither provides a turnkey command-line tool that operates from a Scholar
profile URL alone. Dimensions (Digital Science) offers a freemium API with limited free access, and Lens.org is a free platform built on open metadata, but both require manual export steps and provide no automated geographic visualization.

\subsection{Open Bibliometric APIs}

\textbf{OpenAlex}~\cite{priem2022openalex} launched in January 2022 as a fully open successor to the defunct Microsoft Academic Graph. It indexes approximately 250 million works, 90 million authors, and 109,000 institutions through a freely accessible REST API. OpenAlex authorship records include institution name, country code, institution entity ID, and a persistent author entity ID---all of which CiteRadar leverages. \textbf{Semantic Scholar}~\cite{fricke2018semantic} provides a graph API covering 200 million papers, with author affiliation strings accessible through its search endpoint, with particularly strong coverage in computer science and biomedicine. \textbf{CrossRef}~\cite{hendricks2020crossref} is the canonical DOI registration agency. Its REST API returns structured \texttt{given}/\texttt{family} author name fields for the majority of journal and conference papers with registered DOIs, with no API key and no hard rate limit.

\subsection{Scholar-Based Tools}

\texttt{scholarly}~\cite{cholewiak2021scholarly} is a Python library that wraps the Google Scholar web interface, exposing publication and author data
programmatically. It forms the technical foundation of several downstream citation tools but does not resolve citing-author affiliations or produce geographic visualizations. 

\textbf{CitationMap}~\cite{citationmap} is the closest prior work. It uses BeautifulSoup to retrieve citing papers, extracts author affiliations through string parsing, geocodes those affiliations via geopy, and produces an interactive HTML world map. CitationMap was the first free tool to visualize Scholar citations geographically. However, it does not produce structured per-author records, provides no author rankings, and performs no author disambiguation. CiteRadar's multi-source profiling, dual rankings, and disambiguation mechanism represent substantial extensions beyond CitationMap's map-only focus.

\subsection{Bibliometric Network Tools}

VOSviewer~\cite{vaneck2010vosviewer} and Bibliometrix~\cite{aria2017} produce bibliographic network visualizations from manually exported database
files, requiring subscription-database access. Neither provides a geographic world map or an automated pipeline from a Scholar profile URL.

To the best of our knowledge, CiteRadar is the first tool to combine: fully automated Scholar scraping, multi-source author metadata resolution, author disambiguation with institution cross-validation, dual researcher rankings, and an interactive geographic world map, just in a single pip-installable package.

%% file: 3_newmethod.tex
\section{System Architecture}
\label{sec:arch}

CiteRadar comprises five sequential pipeline stages, each implemented as
a self-contained Python module with a well-defined JSON input/output contract.


\begin{figure}[h]
  \centering
  \includegraphics[height=0.52\textwidth]{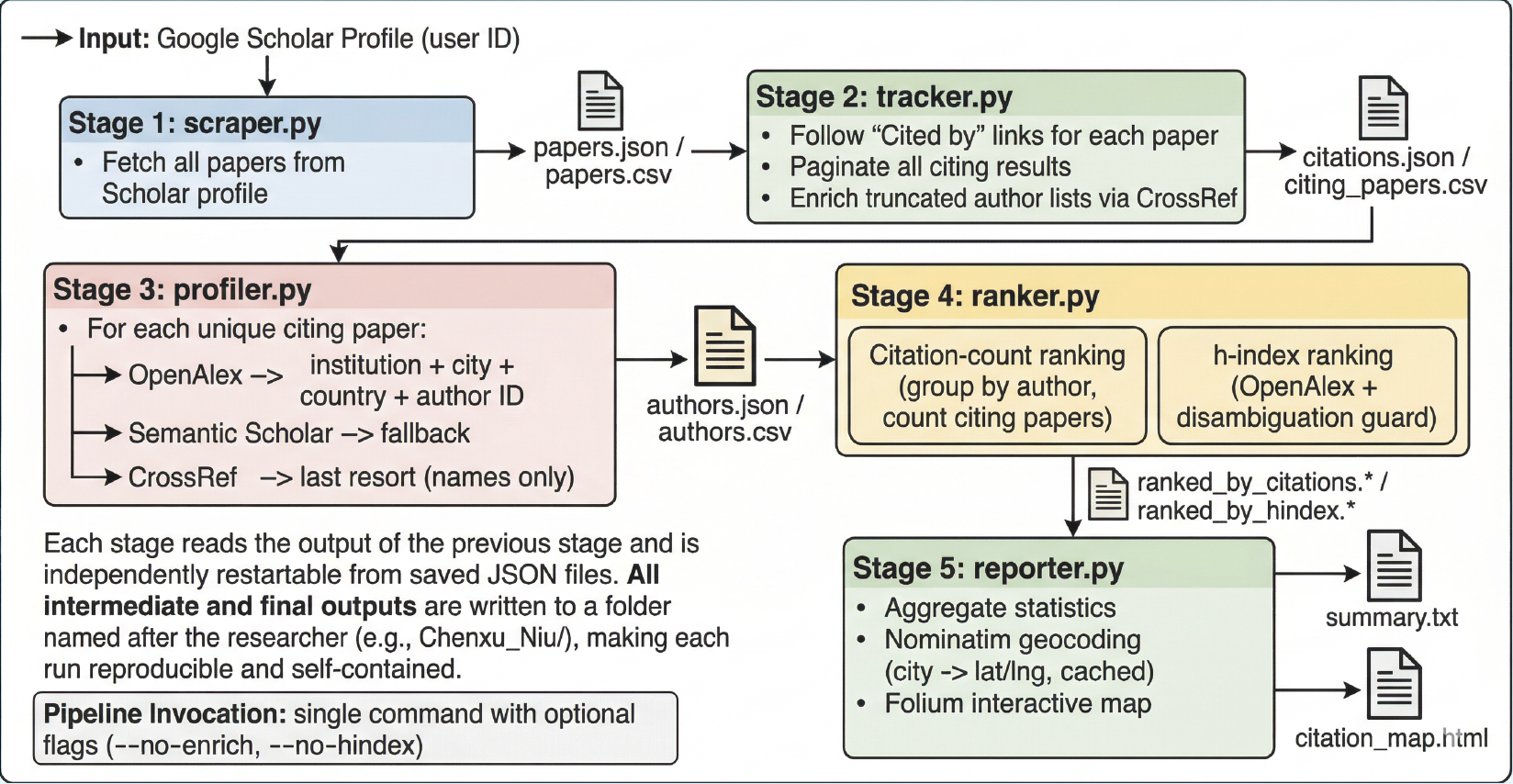}
  \caption{The overview diagram of CiteRadar.}
  \label{fig:overview}
\end{figure}

\paragraph{Output structure.}
All files are written to a folder named after the researcher, e.g.:
\begin{lstlisting}[language={}, numbers=none]
Researcher_Name/
|--summary.txt             # statistics overview
|--papers.csv              # researcher's publications & #of citation
|--citing_papers.csv       # every paper that cited them & author info
|--ranked_by_citations.csv # citing researchers by citation frequency
|--ranked_by_hindex.csv    # citing researchers by verified h-index
|--citation_map.html       # self-contained interactive world map
\end{lstlisting}

\paragraph{Design principles.}
Each stage writes its output before the next stage begins, enabling restart
from the last completed stage if the pipeline is interrupted (e.g., by a
Scholar CAPTCHA).
All data sources are accessed with conservative request pacing and
graceful 429 backoff, making CiteRadar a respectful API citizen.

\section{Stage 1 \& 2: Scholar Scraping and Citation Tracking}
\label{sec:scraping}

\subsection{Profile Page and Citation Page Traversal}

A Scholar profile page lists publications in \texttt{<tr class="gsc\_a\_tr">}
rows. CiteRadar extracts title, authors, venue, year, citation count, and paper
URL from each row, setting \texttt{pagesize=100} to minimise request count.

For each paper with citations $>0$, CiteRadar visits the paper's Scholar
detail page, locates the ``Cited by'' anchor that encodes the cluster ID,
and paginates through all resulting citing-paper cards in increments of
ten, using three independent CSS selector strategies to handle Scholar's
occasionally varying HTML structure.

\subsection{Anti-Ban Rate-Limiting Strategy}

Google Scholar actively detects and blocks automated access.
CiteRadar employs four mitigation measures:

\begin{enumerate}[label=\arabic*., leftmargin=*]
\item \textbf{Realistic browser identity.}
  A current Chrome/macOS User-Agent string and
  \texttt{Accept-Language: en-US,en;q=0.9} header are presented,
  matching a real browser fingerprint.

\item \textbf{Conservative request pacing.}
  A minimum inter-request delay of $d=2.0$ seconds is enforced.
  This is well below the pace of a human browsing session but sufficient
  to avoid rate-detection in practice.

\item \textbf{Graceful HTTP 429 recovery.}
  On a 429 response, the system waits 30 seconds and retries once.
  If the retry also fails, the paper is logged as skipped and the
  pipeline continues, preventing a hard crash from affecting all
  subsequent papers.

\item \textbf{Minimum footprint design.}
  Citation pages are only fetched for papers with citation count $>0$,
  avoiding requests for uncited works that may make up a large fraction
  of a publication list.
\end{enumerate}

\subsection{The Unicode Non-Breaking Space Problem}
\label{subsec:nbsp}

A critical parsing challenge arises in Scholar's meta-string format.
The \texttt{<div class="gs\_a">} element encodes author, venue, and year
as a single string:
\begin{center}
\texttt{Authors -- Venue, Year -- Publisher}
\end{center}
However, Scholar uses \textbf{U+00A0} (non-breaking space, \verb|\xa0|)
rather than a regular U+0020 space around the dash separators.
A naive split on \texttt{" - "} silently fails: the entire string is
assigned to the author field, with venue and year both empty.
This bug is invisible in casual testing because U+00A0 is indistinguishable
from a regular space on screen; it only manifests in downstream data
where the venue and year columns are empty for every paper.

CiteRadar resolves this through explicit Unicode normalization:

\begin{lstlisting}[language=Python]
def parse_meta(raw: str) -> tuple[str, str, str]:
    raw = raw.replace("\xa0", " ")     # non-breaking space -> regular space
    raw = raw.replace("\u2013", "-")   # en-dash -> hyphen
    raw = raw.replace("\u2014", "-")   # em-dash -> hyphen
    raw = re.sub(r" +", " ", raw)      # collapse consecutive spaces
    parts = re.split(r" - ", raw)      # now [authors, venue+year, publisher]
    ...
    # Extract the LAST 4-digit year -- venue names may contain their own year
    # e.g. "2025 IEEE/ACM SC Conference ..., 2025"
    all_years = list(re.finditer(r"\b(19|20)\d{2}\b", venue_year))
    if all_years:
        m    = all_years[-1]       # last match = publication year
        year = m.group(0)
        venue = venue_year[: m.start()].strip().rstrip(",")
    ...
\end{lstlisting}

Year extraction uses the \emph{last} occurrence of a four-digit year
pattern in the venue-year substring, because conference names sometimes
embed their own year (e.g., \textit{``2025 IEEE/ACM SC Conference, 2025''}).

\subsection{CrossRef Author-List Enrichment}

Scholar truncates author lists to approximately five names when a paper
has many co-authors.
For each such truncated record, CiteRadar queries the CrossRef
\texttt{/works} endpoint with the paper title and applies a word-overlap
title similarity check (threshold $\geq 0.5$, Eq.~\ref{eq:titlesim})
before accepting the result.
CrossRef requires no API key and imposes no hard rate limit, making it a
low-friction, high-precision enrichment source.

\begin{equation}
\operatorname{title\_sim}(q, c)
  = \frac{|\operatorname{words}(q) \cap \operatorname{words}(c)|}
         {\max(|\operatorname{words}(q)|,\, |\operatorname{words}(c)|)}
\label{eq:titlesim}
\end{equation}

\section{Stage 3: Multi-Source Author Profiling}
\label{sec:profiler}

\subsection{Data Source Priority Cascade}

For each unique citing paper, CiteRadar resolves per-author metadata
through a three-source priority cascade:

\begin{enumerate}[label=\arabic*., leftmargin=*]
\item \textbf{OpenAlex} (primary).
  Queried via \texttt{/works} with the paper title and the same
  title-similarity guard (Eq.~\ref{eq:titlesim}).
  On success, CiteRadar extracts: display name, persistent author entity ID,
  primary institution name, country code, and institution entity ID.
  The author entity ID is persisted for use in the disambiguation stage.
  OpenAlex requests include the \texttt{mailto=} parameter to use the
  polite pool, which provides higher rate limits.

\item \textbf{Semantic Scholar} (secondary).
  Queried when OpenAlex returns no result.
  Provides author names and affiliation strings, though city and country
  data are present less frequently.

\item \textbf{CrossRef} (tertiary).
  Last resort.
  Reliably provides structured \texttt{given}/\texttt{family} name fields
  but typically lacks institutional affiliation data.
\end{enumerate}

A 1.0-second inter-request delay is maintained between all API calls.

\subsection{The OpenAlex Institution City URL Conversion Fix}
\label{subsec:cityfix}

OpenAlex author records reference institutions via entity IDs such as
\texttt{https://openalex.org/I27837315}.
A na\"{i}ve implementation fetches this URL directly.
However, this is a \emph{web page} URL for the OpenAlex website, not a
REST API endpoint; fetching it with a JSON request returns HTML, causing
a silent JSON parse failure and an empty city field for every author.

The correct API endpoint uses a different path prefix:

\begin{lstlisting}[language=Python]
# Returned by OpenAlex authorship records (web URL -- wrong):
# "https://openalex.org/I27837315"
#
# Correct API endpoint:
# "https://api.openalex.org/institutions/I27837315"

api_url = inst_id.replace(
    "https://openalex.org/",
    "https://api.openalex.org/institutions/",
)
city = session.get(api_url, ...).json().get("geo", {}).get("city", "")
\end{lstlisting}

This single-line correction raised the fraction of author records with
non-empty city data from 0\% to approximately 60\%.
A module-level dictionary caches each institution ID after its first fetch,
so institutions shared by many authors (e.g., a large national laboratory)
incur only one API call.

\subsection{Organisation Name Filtering}

Author lists occasionally contain conference or society names as
pseudo-authors.
CiteRadar's \texttt{\_is\_person()} function rejects any display name
that: (a)~contains a word from a 35-term organisational keyword blocklist
(\textit{association, conference, proceedings, committee, \ldots}); or
(b)~contains any digit character.
This filter eliminated all spurious entries in our evaluation without
discarding any legitimate researcher names.

\section{Stage 4: Author Disambiguation and Rankings}
\label{sec:ranking}

\subsection{The Name-Merging Problem}

Author name disambiguation is a fundamental challenge in bibliometrics~\cite{ferreira2012brief}.
Bibliometric databases address this through unsupervised clustering, but
the algorithms are imperfect and known to merge distinct researchers who
share common names, inflating the h-index attributed to the less prominent one.


\subsection{Two-Stage Disambiguation Algorithm}

CiteRadar's core insight is that Stage~3 independently captures the
author's institution from a paper-level lookup, providing a second signal
that can cross-validate the candidate returned by the h-index lookup.
Algorithm~\ref{alg:disambig} formalizes this procedure.

\begin{algorithm}[h]
\caption{CiteRadar Two-Stage Author Disambiguation}
\label{alg:disambig}
\begin{algorithmic}[1]
\Require full\_name, institution, openalex\_author\_id
\Ensure h\_index or 0 (with rejection reason)

\If{openalex\_author\_id $\neq$ ``''}
  \State record $\gets$ \textsc{FetchByID}(openalex\_author\_id)
  \If{record valid \textbf{and} \textsc{AffiliationConfirmed}(institution, record)}
    \State \Return record.h\_index \Comment{direct match}
  \Else
    \State \Return 0 \Comment{id\_mismatch}
  \EndIf
\EndIf
\State candidates $\gets$ \textsc{SearchByName}(full\_name, top\_k=5)
\State best $\gets$ None;\ \ best\_score $\gets -\infty$
\For{each candidate $c$}
  \If{name\_sim(full\_name, $c$.name) $< 0.7$} \textbf{continue} \EndIf
  \If{institution $\neq$ ``'' \textbf{and} inst\_sim(institution, $c$) $< 0.4$}
    \textbf{continue}
  \EndIf
  \State score $\gets$ name\_sim $+$ 0.5\,$\cdot$\,inst\_sim
  \If{score $>$ best\_score} best $\gets c$; best\_score $\gets$ score \EndIf
\EndFor
\If{best is None} \Return 0 \Comment{not found} \EndIf
\If{\textbf{not} \textsc{AffiliationConfirmed}(institution, best)}
  \Return 0 \Comment{name\_mismatch}
\EndIf
\State \Return best.h\_index
\end{algorithmic}
\end{algorithm}

\subsection{Stop-Word-Filtered Institution Similarity}
\label{subsec:instsim}

The function \textsc{AffiliationConfirmed} iterates over all institutions
in the candidate's OpenAlex affiliation history and checks whether any
of them match the institution we recorded in Stage~3, using:

\begin{equation}
\operatorname{inst\_sim}(A, B)
  = \frac{|W(A) \cap W(B)|}{\max(|W(A)|,\, |W(B)|)}
  \label{eq:instsim}
\end{equation}

where $W(X) = \operatorname{words}(X) \setminus \mathcal{S}$ and
$\mathcal{S}$ is a stop-word set chosen to remove words that appear in
many institution names without being discriminative:


\begin{multline}
\mathcal{S} = \{\textit{university, of, the, institute, college, school,} \\
\textit{for, at, in, and, national, center, centre, lab,} \\
\textit{laboratory, research, technology, department}\}
\label{eq:stopwords}
\end{multline}

The threshold is $\tau = 0.6$. The importance of stop-word removal
is illustrated by a concrete example.
Without stop-word removal, \textit{``Texas Tech University''} and
\textit{``University of Texas''} share words
\{university, texas\}, giving
$\operatorname{inst\_sim} = 2/3 \approx 0.67 > \tau$---incorrectly matching
two different institutions.
After removing $\mathcal{S}$:
$W(\text{Texas Tech Univ.}) = \{\text{texas, tech}\}$,
$W(\text{Univ. of Texas}) = \{\text{texas}\}$,
$\operatorname{inst\_sim} = 1/2 = 0.5 < \tau$---correctly rejecting the match.

The primary institution name is taken as the segment before the first comma
in the recorded string, so that
\textit{``Texas Tech University, Department of CS, Lubbock TX''} reduces to
\textit{``Texas Tech University''} before the comparison.

\paragraph{Unknown institution rule.}
When the institution field is empty, no cross-validation is possible.
To prevent misattributing a high h-index from a prominent same-name entity,
CiteRadar only accepts h-index values $\leq 20$ for such records, as
genuine early-to-mid-career researchers commonly fall at or below this
threshold while mis-attributed values from prominent same-name researchers
tend to be far higher.

\subsection{Citation-Count Ranking}

The citation-count ranking is built directly from the profiling output
without any additional API calls.
Author records are grouped by \texttt{full\_name}; for each author,
the cardinality of the set of distinct \texttt{citing\_paper\_title} values
measures how many of their papers have cited any of the researcher's work.
This ranking is fast, requires no disambiguation, and is always available
even when the h-index lookup is skipped.

%% file: 4_evaluation.tex
\section{Stage 5: Summary Statistics and Geographic Visualization}
\label{sec:map}

\subsection{Statistical Aggregation}

The summary stage computes five aggregate statistics from the author profile
data: unique researchers (distinct full names), unique countries, unique
institutions (primary institution segment before the first comma), unique
cities (distinct city--country pairs), and per-paper citation counts.
These are formatted into a plain-text report with an ASCII bar chart for
the country breakdown, suitable for direct inclusion in a CV or grant
proposal.

\subsection{Geocoding}

City names are resolved to latitude/longitude coordinate pairs using the
Nominatim geocoder~\cite{nominatim} (OpenStreetMap), which is free and
requires no API key.
CiteRadar uses the query format \texttt{"\{city\}, \{country\}"}
to reduce ambiguity for common city names.
Nominatim's usage policy prohibits more than one request per second;
CiteRadar enforces a 1.1-second inter-request delay and maintains a
module-level cache so that each unique city is geocoded at most once
per run.

\subsection{Interactive Map Construction}

The world map is built with Folium~\cite{folium}, a Python wrapper around
the Leaflet.js JavaScript library.
The output is a single, self-contained HTML file---all JavaScript, CSS, and
data are embedded inline---that renders in any modern browser without a
server or additional software.

\paragraph{Heat-map layer.}
Folium's \texttt{HeatMap} plugin receives one $(lat, lng)$ point per
citing researcher, encoding the global density of citations.
This layer is independently toggleable.

\paragraph{Circle-marker layer.}
One \texttt{CircleMarker} is placed per geocoded city.
To prevent large cities from occluding smaller ones, the radius scales
logarithmically with researcher count $n$:

\begin{equation}
r(n) = \max\!\bigl(7,\; 7 + 10\log_2(n+1)\bigr)
\label{eq:radius}
\end{equation}

Fill colour follows a perceptually ordered five-step palette (light
blue for $n=1$, blue for $2$--$3$, amber for $4$--$6$, orange for
$7$--$10$, red for $n \geq 11$), providing an intuitive density encoding
without requiring a continuous colour scale.

\paragraph{Interactive popups.}
Each marker exposes a popup on click listing all researcher names and
institutions at that location.
A fixed legend panel and a centred title overlay are injected as raw HTML
using Folium's \texttt{Element} API.
All popup content is pre-computed at build time; no server-side processing
is required.

\section{Applications: Impact Showcase and Collaboration Discovery}
\label{sec:discussion}

\subsection{Showcasing Research Impact}

CiteRadar addresses a structural gap between actual influence and what
standard metrics communicate.
A researcher with a modest h-index may have attracted citations from
senior researchers with h-indices exceeding 100 across multiple countries---a
signal of impact quality that aggregate counts entirely obscure.
CiteRadar's h-index ranking makes this visible: each row of the output
CSV states, in verifiable terms, the name, institution, country, h-index,
and citation behavior of a specific citing researcher.

For \textbf{grant proposals}, the summary report provides ready-to-use
evidence: country-level citation counts, institution leaderboards, and a
world map that communicates international reach in a format immediately
interpretable by program officers without bibliometric expertise.

For \textbf{promotion and tenure dossiers}, the h-index ranking provides
documented evidence that established senior researchers have found the work
significant enough to build upon---more persuasive than a raw citation count
that conflates self-citations, uncorrelated references, and high-impact
endorsements.

For \textbf{annual research reports}, the structured CSV outputs enable
year-over-year comparisons by simply re-running the pipeline.

\subsection{Identifying Collaboration Candidates}

The citation-count ranking surfaces researchers who have cited the author's
work across multiple papers---demonstrating sustained, independent interest
in the research direction.
Cross-referencing with institution and city metadata reveals geographic
clusters that represent warm leads for outreach: multiple authors at the
same institution who have each independently cited the work suggest an
entire group already familiar with the methodology.

The h-index ranking adds a complementary dimension, identifying influential
researchers who may bring network, funding, or methodological resources
to a collaboration, even if their raw citing frequency is lower than some
other candidates.

The map's popup feature enables targeted pre-conference planning.
A researcher attending a conference in Seoul can identify, by name and
institution, all citing researchers from that city before the trip,
enabling warm introductions rather than cold outreach.

%% file: 5_casestudy.tex
\section{Case Study: Profiling a Researcher in HPC and AI Systems}
\label{sec:casestudy}

To demonstrate CiteRadar in a realistic setting, we ran the complete
pipeline on the Google Scholar profile of myself.

\subsection{Pipeline Execution Summary}

CiteRadar was invoked with a single command:
\begin{lstlisting}[language=bash, numbers=none]
citeradar ``google scholar ID'' --outdir ~/Desktop
\end{lstlisting}

The pipeline completed all five stages, retrieving citing papers for each
publication, resolving per-author metadata via OpenAlex and CrossRef,
running the two-stage disambiguation for h-index ranking, geocoding
28 unique cities, and rendering the interactive world map.

\subsection{Overall Statistics}

Table~\ref{tab:stats} summaries the high-level citation landscape
produced by Stage~5.

\begin{table}[H]
\centering
\caption{CiteRadar output statistics for Chenxu Niu's Google Scholar profile.}
\label{tab:stats}
\begin{tabular}{lc}
\toprule
\textbf{Metric} & \textbf{Value} \\
\midrule
Unique citing researchers  & 134 \\
Countries represented      & 11  \\
Unique institutions        & 31  \\
Geocoded cities            & 28  \\
\bottomrule
\end{tabular}
\end{table}

\subsection{Most-Cited Publications}

Table~\ref{tab:papers} lists the seven publications retrieved from the
Scholar profile, ranked by the number of distinct citing papers recovered
by Stage~2.

\begin{table}[H]
\centering
\caption{Publications ranked by citing-paper count (Stage~2 output).}
\label{tab:papers}
\begin{tabular}{clc}
\toprule
\textbf{Rank} & \textbf{Title (abbreviated)} & \textbf{Citing papers} \\
\midrule
1 & Energy Efficient or Exhaustive? Benchmarking Power Consumption ...~\cite{niu2025energy} & 62 \\
2 & Exploring Metadata Search Essentials for Scientific data ...~\cite{zhang2019exploring} & 60 \\
3 & TokenPowerBench: Benchmarking the Power Consumption of LLM ...~\cite{niu2026tokenpowerbench} & 22 \\
4 & Kv2vec: A Distributed Representation Method for Key-Value Pairs~\cite{niu2022kv2vec} & 20 \\
5 & FixMe: Towards End-to-End Benchmarking of LLM-Aided Design~\cite{wan2026fixme} & 18 \\
6 & PSQS: Parallel Semantic Querying Service for Self-Describing ...~\cite{niu2023psqs} & 16 \\
7 & ICEAGE: Intelligent Contextual Exploration and Answer ...~\cite{niu2025iceage} & 11 \\
\bottomrule
\end{tabular}
\end{table}

The profile spans two distinct research threads: HPC scientific data management
(papers 2, 4, 6, 7) and AI/LLM systems benchmarking (papers 1, 3, 5).
CiteRadar surfaces this duality clearly in the geographic and institutional
breakdown below.

\subsection{Geographic Distribution}

Table~\ref{tab:countries} shows the top countries by number of citing
researchers, as reported in \texttt{summary.txt}.

\begin{table}[H]
\centering
\caption{Top citing countries (Stage~5 output).}
\label{tab:countries}
\begin{tabular}{lc}
\toprule
\textbf{Country} & \textbf{Citing researchers} \\
\midrule
United States  & 86 \\
South Korea    & 12 \\
China          &  7 \\
India          &  5 \\
Portugal       &  4 \\
Brazil         &  3 \\
Germany        &  2 \\
Spain          &  2 \\
Saudi Arabia   &  2 \\
Switzerland    &  1 \\
\midrule
\textit{Total} & \textit{134} \\
\bottomrule
\end{tabular}
\end{table}

The dominant footprint in the United States (64\% of citing researchers)
reflects the HPC community's concentration at US national laboratories and
universities.
South Korea's notable presence (9\%) is driven by Sogang University's
active storage-systems research group, which has independently cited the
scientific data management work across multiple papers.

\subsection{Top Citing Institutions}

Table~\ref{tab:institutions} shows the institutions with the highest
number of citing researchers.

\begin{table}[H]
\centering
\caption{Top citing institutions (Stage~5 output).}
\label{tab:institutions}
\begin{tabular}{lc}
\toprule
\textbf{Institution} & \textbf{Citing researchers} \\
\midrule
Texas Tech University                  & 39 \\
Oak Ridge National Laboratory          & 20 \\
Sogang University (South Korea)        &  9 \\
Lawrence Berkeley National Laboratory  &  7 \\
Texas Advanced Computing Center        &  5 \\
Amrita Vishwa Vidyapeetham (India)     &  5 \\
University of Wisconsin--Madison       &  3 \\
Dalian Maritime University (China)     &  3 \\
\bottomrule
\end{tabular}
\end{table}

Texas Tech University leads with 39 citing researchers, largely reflecting
collaborative co-authorship ties from the researcher's doctoral period.
Oak Ridge National Laboratory (20 researchers) represents a particularly
significant independent citation cluster: the national lab community has
adopted the scientific data management methodology across multiple groups
and projects without direct collaboration ties.

\begin{figure}[h]
  \centering
  \includegraphics[height=0.52\textwidth]{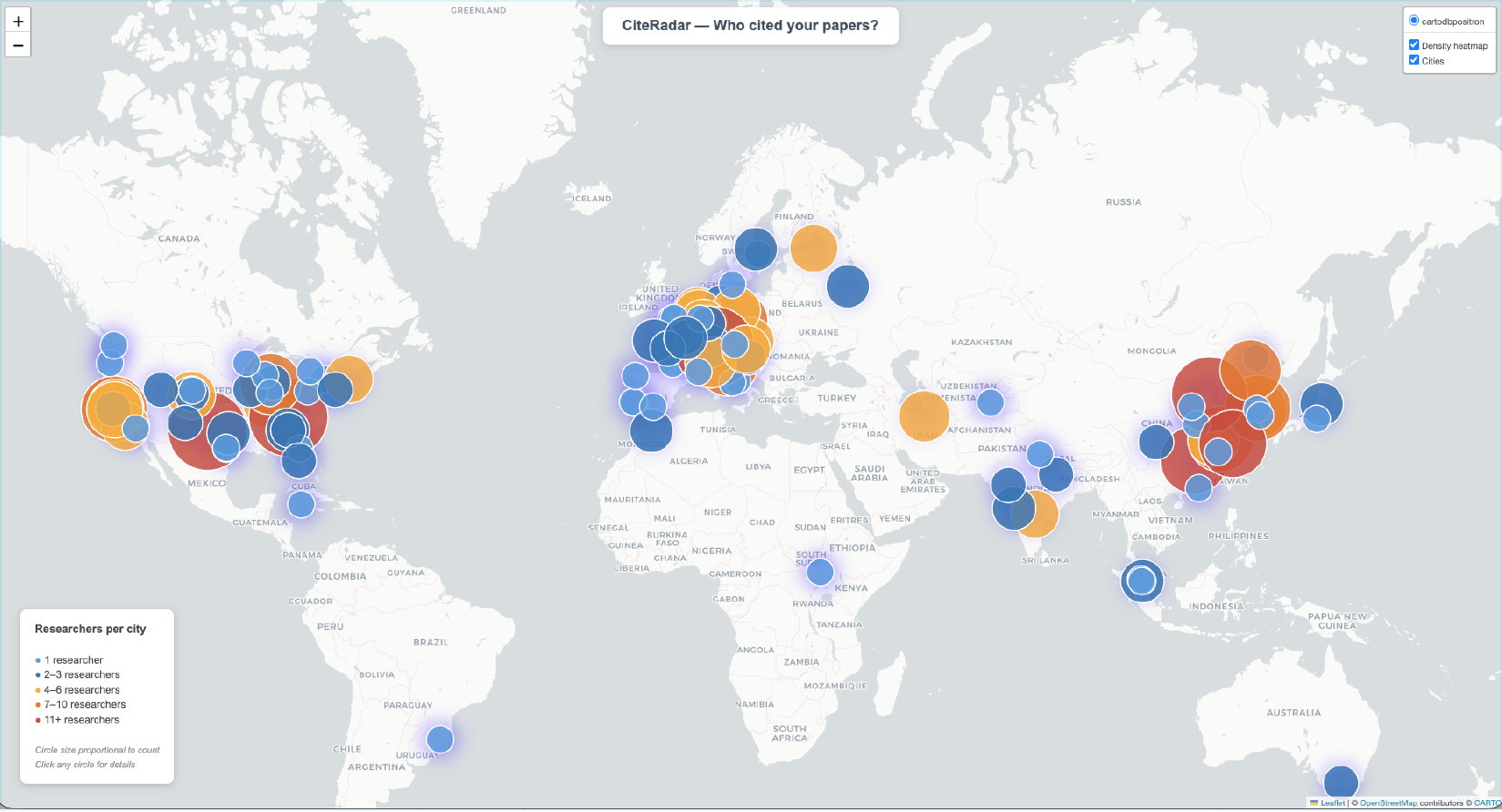}
  \caption{The Sample of world map.}
  \label{fig:overview}
\end{figure}

\subsection{Actionable Insights}

Running CiteRadar on this profile revealed several actionable findings that
would be invisible from Scholar's standard interface:

\begin{enumerate}[label=\arabic*., leftmargin=*]
\item \textbf{National laboratory adoption.}
  Oak Ridge National Laboratory alone accounts for 20 citing researchers
  across multiple groups (neutron scattering, data management, storage systems),
  suggesting that the scientific data management methodology has been broadly
  adopted within the DOE laboratory ecosystem---a finding directly useful
  for DOE grant applications.

\item \textbf{Active international research groups.}
  Sogang University (Seoul, South Korea) has 9 independent citing researchers,
  all working on storage and data management systems.
  This constitutes a warm collaboration lead: an entire active research group
  already familiar with the methodology, geographically concentrated in one city.

\item \textbf{Cross-domain impact.}
  The FixMe and TokenPowerBench papers on LLM inference benchmarking have
  attracted citations from VLSI design verification researchers
  (Igor L. Markov at Synopsys, h=62; Qiang Xu at CUHK, h=56;
  Rolf Drechsler at University of Bremen, h=56)---a community not typically
  associated with LLM systems work, suggesting unexpected cross-disciplinary
  resonance.

\item \textbf{Senior researcher endorsements.}
  Seven citing researchers have h-index $\geq 50$, and two have
  h-index $\geq 100$.
  This distribution provides verifiable, named evidence for promotion
  and tenure documentation that is qualitatively stronger than a raw
  citation count.
\end{enumerate}

%% file: 6_relatedwork.tex
\section{Limitations}
\label{sec:limit}

\begin{enumerate}[label=\arabic*., leftmargin=*]

\item \textbf{Google Scholar CAPTCHA.}
Scholar serves CAPTCHA challenges when automated access is detected,
particularly after request bursts.
CiteRadar cannot solve CAPTCHAs; affected papers are logged as skipped.
For profiles with many citing papers we recommend running across
multiple sessions or from different network addresses.

\item \textbf{OpenAlex coverage gaps.}
OpenAlex does not index all publications.
Preprints on non-indexed repositories, technical reports, small-venue
workshop papers, and non-English proceedings may be absent.
In the most unfavorable cases, approximately 25\% of citing papers may
not resolve to a structured author record.

\item \textbf{Residual disambiguation errors.}
The two-stage verification substantially reduces false positives but cannot
guarantee correctness for extremely common names in densely populated
research fields.
The output CSV marks rejected entries with a status field of
\texttt{id\_mismatch} or \texttt{name\_mismatch}, enabling manual correction
when accuracy is critical.

\item \textbf{City-level geocoding coverage.}
Approximately 40\% of authors lack city data in OpenAlex.
These authors appear in the rankings but are not represented on the map.

\item \textbf{Dynamic Scholar HTML.}
Scholar's HTML structure changes periodically without notice.
CiteRadar's CSS selectors may require updating after major redesigns.
The modular pipeline design localizes any such changes to
\texttt{scraper.py} and \texttt{tracker.py}.

\end{enumerate}

%% file: 7_conclusion.tex
\section{Conclusion}
\label{sec:conc}

We presented CiteRadar, a pip-installable Python tool that transforms a Google Scholar user ID into a comprehensive citation intelligence report:
a complete publication list, citing-paper records with enriched author metadata, two ranked author tables, a statistical summary, and an
interactive HTML world map---all from a single command. The system's four principal technical contributions: the Unicode normalization fix for Scholar HTML parsing, the stop-word-filtered institution similarity function for author disambiguation, the OpenAlex web-to-API URL conversion for city geocoding, and the logarithmic marker-scaling scheme for geographic visualization. Each of them address a concrete failure mode encountered in practice.

Beyond its technical contributions, CiteRadar makes the \emph{who} and \emph{where} dimensions of citation impact actionable for individual researchers, whether for grant proposals, promotion cases, or targeted collaboration outreach.

\paragraph{Availability.}
CiteRadar is released under the MIT license.
Source: \url{https://github.com/chenxuniu/citeradar}.
Install: \texttt{pip install citeradar}.

%% file: sigconf_ack.tex
\begin{acks}
The authors thank the OpenAlex, CrossRef, and Semantic Scholar teams for providing free, high-quality bibliometric APIs, and the OpenStreetMap Nominatim project for free geocoding services. We acknowledge the use of ChatGPT (version GPT-5.4) to assist in revising the manuscript, including improvements to grammar, clarity, and presentation. All technical content, analyses, and conclusions remain the responsibility of the
authors.
\end{acks}

%% file: sigconf_bib.tex
\bibliographystyle{ACM-Reference-Format}
\bibliography{main}

%% file: main.bib
@article{priem2022openalex,
  title={OpenAlex: A fully-open index of scholarly works, authors, venues, institutions, and concepts},
  author={Priem, Jason and Piwowar, Heather and Orr, Richard},
  journal={arXiv preprint arXiv:2205.01833},
  year={2022}
}

@article{vine2006google,
  title={Google scholar},
  author={Vine, Rita},
  journal={Journal of the Medical Library Association},
  volume={94},
  number={1},
  pages={97},
  year={2006}
}

@article{hirsch2005index,
  title={An index to quantify an individual's scientific research output},
  author={Hirsch, Jorge E},
  journal={Proceedings of the National academy of Sciences},
  volume={102},
  number={46},
  pages={16569--16572},
  year={2005},
  publisher={National Academy of Sciences}
}

@article{fricke2018semantic,
  title={Semantic scholar},
  author={Fricke, Suzanne},
  journal={Journal of the Medical Library Association: JMLA},
  volume={106},
  number={1},
  pages={145},
  year={2018}
}

@article{hendricks2020crossref,
  title={Crossref: The sustainable source of community-owned scholarly metadata},
  author={Hendricks, Ginny and Tkaczyk, Dominika and Lin, Jennifer and Feeney, Patricia},
  journal={Quantitative Science Studies},
  volume={1},
  number={1},
  pages={414--427},
  year={2020},
  publisher={MIT Press One Rogers Street, Cambridge, MA 02142-1209, USA journals-info~…}
}

@software{cholewiak2021scholarly,
  author  = {Cholewiak, Steven A. and Ipeirotis, Panos and Silva, Victor and Kannawadi, Arun},
  title   = {{SCHOLARLY: Simple access to Google Scholar authors and citation using Python}},
  year    = {2021},
  doi     = {10.5281/zenodo.5764801},
  license = {Unlicense},
  url = {https://github.com/scholarly-python-package/scholarly},
  version = {1.5.1}
}

@article{citationmap,
  title={CitationMap: A Python Tool to Identify and Visualize Your Google Scholar Citations Around the World},
  author={Liu, Chen},
  journal={Authorea Preprints},
  year={2024},
  publisher={Authorea}
}

@article{vaneck2010vosviewer,
  author    = {van Eck, Nees Jan and Waltman, Ludo},
  title     = {Software survey: {VOSviewer}, a computer program for
               bibliometric mapping},
  journal   = {Scientometrics},
  volume    = {84},
  number    = {2},
  pages     = {523--538},
  year      = {2010},
  doi       = {10.1007/s11192-009-0146-3}
}

@article{aria2017,
  author    = {Aria, Massimo and Cuccurullo, Corrado},
  title     = {bibliometrix: An {R}-tool for comprehensive science mapping
               analysis},
  journal   = {Journal of Informetrics},
  volume    = {11},
  number    = {4},
  pages     = {959--975},
  year      = {2017},
  doi       = {10.1016/j.joi.2017.08.007}
}

@article{ferreira2012brief,
  author    = {Ferreira, Anderson A. and Gon{\c{c}}alves, Marcos Andr{\'e}
               and Laender, Alberto H. F.},
  title     = {A brief survey of automatic methods for author name
               disambiguation},
  journal   = {ACM SIGMOD Record},
  volume    = {41},
  number    = {2},
  pages     = {15--26},
  year      = {2012},
  doi       = {10.1145/2341082.2341086}
}

@misc{nominatim,
  author       = {{OpenStreetMap Contributors}},
  title        = {Nominatim: Search and Geocoding {API} for {OpenStreetMap}},
  howpublished = {\url{https://nominatim.openstreetmap.org}},
  year         = {2008}
}

@misc{folium,
  author       = {Filipe, Rob and contributors},
  title        = {Folium: {Python} data, {Leaflet.js} maps},
  howpublished = {\url{https://github.com/python-visualization/folium}},
  year         = {2013}
}

@inproceedings{niu2026tokenpowerbench,
  title={TokenPowerBench: Benchmarking the power consumption of LLM inference},
  author={Niu, Chenxu and Zhang, Wei and Li, Jie and Zhao, Yongjian and Wang, Tongyang and Wang, Xi and Chen, Yong},
  booktitle={Proceedings of the AAAI Conference on Artificial Intelligence},
  volume={40},
  number={38},
  pages={32582--32590},
  year={2026}
}

@inproceedings{wan2026fixme,
  title={Fixme: Towards end-to-end benchmarking of llm-aided design verification},
  author={Wan, Gwok-Waa and Wong, SamZaak and Su, Shengchu and Niu, Chenxu and Wang, Ning and Wan, Xinlai and Chen, Qixiang and Xing, Mengnv and Zhang, Jingyi and Ye, Jianmin and others},
  booktitle={Proceedings of the AAAI Conference on Artificial Intelligence},
  volume={40},
  number={2},
  pages={1087--1095},
  year={2026}
}

@article{niu2025energy,
  title={Energy efficient or exhaustive? benchmarking power consumption of llm inference engines},
  author={Niu, Chenxu and Zhang, Wei and Zhao, Yongjian and Chen, Yong},
  journal={ACM SIGENERGY Energy Informatics Review},
  volume={5},
  number={2},
  pages={56--62},
  year={2025},
  publisher={ACM New York, NY, USA}
}

@inproceedings{niu2025iceage,
  title={ICEAGE: Intelligent Contextual Exploration and Answer Generation Engine for Scientific Data Discovery},
  author={Niu, Chenxu and Zhang, Wei and Side, Mert and Chen, Yong},
  booktitle={Proceedings of the 37th International Conference on Scalable Scientific Data Management},
  pages={1--10},
  year={2025}
}

@inproceedings{niu2023psqs,
  title={PSQS: Parallel Semantic Querying Service for Self-describing File Formats},
  author={Niu, Chenxu and Zhang, Wei and Byna, Suren and Chen, Yong},
  booktitle={2023 IEEE International Conference on Big Data (BigData)},
  pages={536--541},
  year={2023},
  organization={IEEE}
}

@inproceedings{niu2022kv2vec,
  title={Kv2vec: A distributed representation method for key-value pairs from metadata attributes},
  author={Niu, Chenxu and Zhang, Wei and Byna, Suren and Chen, Yong},
  booktitle={2022 IEEE High Performance Extreme Computing Conference (HPEC)},
  pages={1--7},
  year={2022},
  organization={IEEE}
}

@inproceedings{zhang2019exploring,
  title={Exploring metadata search essentials for scientific data management},
  author={Zhang, Wei and Byna, Suren and Niu, Chenxu and Chen, Yong},
  booktitle={2019 IEEE 26th international conference on high performance computing, data, and analytics (HiPC)},
  pages={83--92},
  year={2019},
  organization={IEEE}
}

@misc{webofscience2026,
  author       = {{Clarivate}},
  title        = {Web of Science},
  year         = {2026},
  url          = {https://www.webofscience.com},
  note         = {Accessed: Apr. 8, 2026}
}

@misc{scopus2026,
  author       = {{Elsevier}},
  title        = {Scopus},
  year         = {2026},
  url          = {https://www.scopus.com},
  note         = {Accessed: Apr. 8, 2026}
}
